%% file: acl_latex.tex
\pdfoutput=1
\documentclass[11pt]{article}

\usepackage[final]{acl}

\usepackage{times}
\usepackage{booktabs}
\usepackage{tabularx}
\usepackage{latexsym}

\usepackage[T1]{fontenc}
\usepackage[utf8]{inputenc}

\usepackage{graphicx}
\usepackage{microtype}

\usepackage{inconsolata}

\usepackage{listings}
\usepackage{xcolor}
\definecolor{myblue}{RGB}{123, 194, 76}

\newcolumntype{b}{X}
\newcolumntype{s}{>{\hsize=.15\hsize}X}
\newcolumntype{m}{>{\hsize=.3\hsize}X}

\title{Using Adaptive Empathetic Responses for Teaching English}

\author{Li Siyan, Teresa Shao, Zhou Yu, Julia Hirschberg\\
Department of Computer Science\\
Columbia University\\
\texttt{\{siyan.li,ts3488,zy2461,jbh2019\}@columbia.edu}}

\begin{document}
\maketitle
\begin{abstract}
Existing English-teaching chatbots rarely incorporate empathy explicitly in their feedback, but empathetic feedback could help keep students engaged and reduce learner anxiety. Toward this end, we propose the task of negative emotion detection via audio, for recognizing empathetic feedback opportunities in language learning. We then build the first spoken English-teaching chatbot with adaptive, empathetic feedback. This feedback is synthesized through automatic prompt optimization of ChatGPT and is evaluated with English learners. We demonstrate the effectiveness of our system through a preliminary user study.
\end{abstract}

\section{Introduction}

Teacher empathy has been shown to improve the learning experience, including increasing learner engagement and reducing anxiety \cite{cooper2002teachers,lam2011empathy,zhang2022toward}.
Recently, \citet{wu2023teachers} suggests that students' perceived affective support (PAS) from teachers has a positive correlation with {\it L2 grit}, defined as the passion and perseverance for second-language learning \cite{teimouri2022l2}. PAS generally corresponds to the perceived level of support for emotional needs. Its definition includes caring, valuing responses, listening, and encouragement \cite{sakiz2007does}. We therefore expect empathy to correlate positively with PAS. We aim to examine whether an empathetic, English-teaching system with high PAS similarly boosts L2 grit.

English-teaching systems have adopted affective components for various purposes \cite{zhai2022systematic}. However, the systems that have introduced empathetic components into pedagogy are either situated in ubiquitous learning environments \cite{dai2014emotional,santos2016toward} or are not naturalistic or seamless in their approaches to accounting for student affect \cite{wu2022affective}. An interactive system that effectively detects and alleviates ESL learner anxiety without sensors (e.g. pulse rate monitors) or cameras has yet to be implemented. 

Detecting negative emotion from a learner's audio is a promising way to offer empathetic feedback. However,  off-the-shelf English speech emotion recognition models are often trained on data collected with native speakers of English \cite{busso2008iemocap,lotfian2017building}. We hypothesize that English spoken by non-native speakers will have differences \cite{lin2014understanding} that challenge these off-the-shelf models. To address this, we develop a preliminary pipeline for this task using annotated audio data and incorporate it into a spoken empathetic chatbot system.

Our spoken chatbot detects negative emotions or prolonged pauses and then responds empathetically to encourage students. This negative-emotion-responsive design is inspired by an automated physics tutor that senses student frustration using sensors and cameras \cite{d2013autotutor}. 
We currently employ model-based and automatic approaches for sensing negative affect in user audio. 
The chatbot also provides grammar feedback. We utilize a grammar correction model for grammatical feedback and ChatGPT with optimized prompting for empathetic feedback. Positive preliminary user study results indicate that users perceive affective support from our system, paving the way for future large-scale experiments to study our system's effect on learner L2 grit.

Our main contributions are: 1) We release a dataset of Mandarin-accented English speech with high-quality ASR transcripts and negative emotion annotations, and 2) We propose the first sensor-free educational English chatbot that detects negative affect and intervenes by providing adaptive empathetic feedback \footnote{The dataset and code are in \url{https://github.com/siyan-sylvia-li/adaptive_empathetic_BEA2024}}.

\section{Related Work}

\subsection{Emotion Recognition in English-Teaching Systems}

Past English-teaching systems often relied on facial emotion recognition for detecting user affective states. \citet{lin2015perspective} features a teaching assistant that recognizes the user's emotional state from facial expressions and then adjusts the material's difficulty. 
\citet{zhang2022affective} proposes a convolutional neural network-based approach to learner emotion recognition to be used in future systems. We are not considering the visual modality due to the constraints of the dialogue framework we build upon. 

\citet{mazur2011proposal} creates a gamified scoring system to adapt to different users. 
This system is equipped with affect classification for Japanese textual input, yet the role of empathy here is unclear.

Other systems have employed less seamless approaches to detecting affect changes. \citet{wu2022affective} constructs an emotion recognition module by recording the number of times a learner clicks on positive and negative emoticons. \citet{santos2016toward} employs Arduino, an open-source electronic prototyping platform to detect learner physiological changes, such as pulses and skin conductivity. 
Another ubiquitous learning system, \citet{dai2014emotional}, uses speech signal and multi-agent behavioral data for online learning and a neural mechanism model for analyzing learners' emotional characteristics. 

\begin{figure*}[btp]
    \centering
    \includegraphics[scale=0.3]{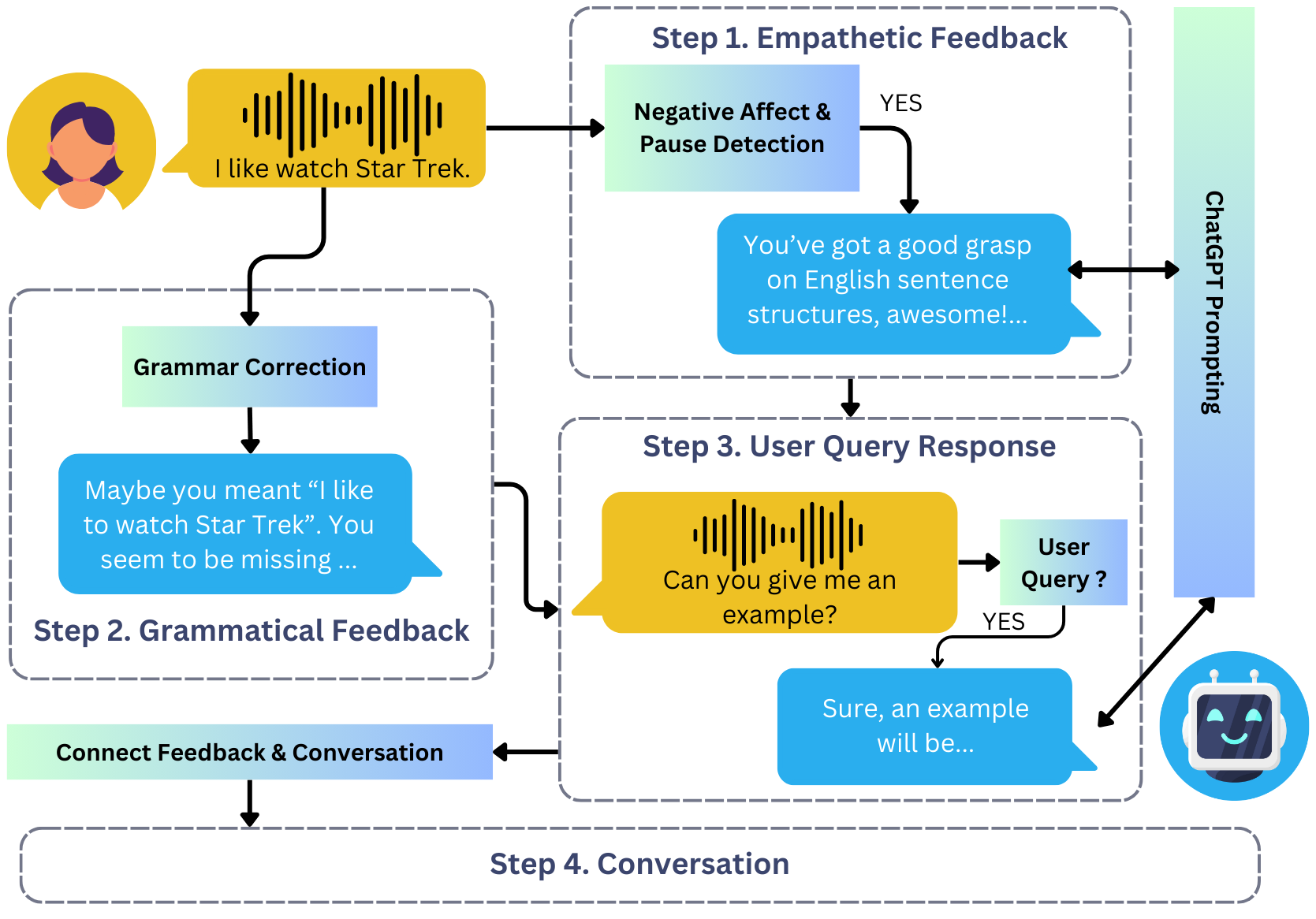}
    \caption{System Design Overview.}
    \label{fig:overview}
\end{figure*}

\subsection{Affective English-Teaching Chatbots}

Chatbots are effective in increasing student conversational activity during discussions \cite{goda2014conversation}, improving listening skills \cite{kim2018study} and grammar \cite{kim2019future}, and enhancing writing abilities \cite{lin2020enhancing}. Since ChatGPT appeared, the quality of chatbot responses has improved dramatically, eliminating concerns about adverse effects on student outcomes due to low response quality \cite{fryer2020bots}. 

\citet{ayedoun2015conversational} introduces a multimodal agent that simulates a restaurant waiter to situate participants in a social conversational context to improve their willingness to communicate. 
\citet{ayedoun2020toward} further improves this system by incorporating communication strategies and affective backchannels to provide personalized scaffolding. 
While the systems alleviate learner anxiety, learner emotions are not directly accounted for or addressed. Both systems also rely on pre-scripted dialogue and are restricted in scenarios.

\citet{shi2020design} builds an empathetic spoken chatbot into a WeChat program for English tutoring. The GPT-2-based \cite{radford2019language} chatbot utilizes an ontology and a retrieval-based generation approach similar to XiaoIce \cite{zhou2020design}. Despite being empathetic, the bot only uses audio for pronunciation correction. 

\subsection{Pauses and Anxiety in ESL Context}
Foreign language anxiety can correlate with higher pause rates and lower fluency. \citet{perez2019role} established that learners with higher language anxiety tend to pause more frequently. In a study by \citet{wilang2018complexity} that monitors ESL speakers speaking during an exam, pausing is associated with heart rate spikes for some, indicating anxiety during pauses. ESL teachers have also noted pauses and stammering as signs of students struggling with language anxiety \cite{kasap2019anxiety}.

\section{System Design}
\subsection{Overview}
Figure~\ref{fig:overview} shows a system overview: User audio is sent to the \textit{Empathetic Feedback} module to determine whether the user is distressed. If so, the bot produces empathetic feedback using past user utterances; otherwise, the system continues to the \textit{Grammatical Feedback} stage, where grammar critiques are given if applicable. If either feedback mechanism is triggered, the system transitions back to the original conversation through the \textit{User Query Response} stage if the user follows up with the feedback, then through the \textit{Connect Feedback \& Conversation} module. To avoid overwhelming users, we ensure at least two turns between grammatical feedback and four turns between empathetic feedback. We discussed our design with ESL students and consulted teachers before finalizing our system. See Appendix \ref{app:discussion} for details. 

We build on an existing dialogue framework \cite{li2022using} for speech and text dialogue system development. The system allows users to converse with the chatbot by recording their utterances through a microphone. The utterance is then converted to text using Whisper medium \cite{radford2023robust} and the text and audio are sent to the chatbot for further analyses and response synthesis. The chatbot response is spoken using SpeechT5 \cite{ao-etal-2022-speecht5}. For the specific speaker embedding, we selected one of the \texttt{slt} clips from the CMU Arctic speech databases\cite{kominek2004cmu} manually. When choosing the speaker embedding, we aimed for a female voice that can sufficiently induce perceived empathy.

\subsection{Data Used}
To create data for testing various modules, we utilized audio clips of native Mandarin speakers conversing with a chatbot collected from an English practice platform \cite{li2022using}. 3,200 audio clips from 613 conversations and 163 users remained after filtering. The filtering process removes audio clips containing only Mandarin, duplicates, and a subset of self-introductions from the users. We were not able to eliminate all identifying information from this stage of filtering, but we will remove all identifiable information before publicizing our data. Each audio clip ranges from one second to two minutes. We did not collect demographic information for user identity protection.

We transcribed all audio clips with Whisper medium for training the text-based models in our pipeline. Whisper is not always sufficiently robust to handle heavily accented speech in our data; however, to realistically simulate the environment for our models, we choose not to correct these transcriptions, although we will release the data after manually correcting the transcripts to ensure quality. Realistically, our system should improve as more accent-robust real-time ASR systems emerge. 

\begin{table*}[!h]
    \centering
    \begin{tabularx}{\linewidth}{XXX}
        \toprule
        \textbf{Input} & \textbf{Correction} & \textbf{Llama}\\
        \midrule
        I like to read book and study English. & I like to read books and study English & I like to read books and study English. I also like to spend time with my friends.\\
        \midrule
        Love story & Love story. & Love story. Maybe I will write a book one of these days. \\
        \bottomrule
    \end{tabularx}
    \caption{Examples of the trained Llama model extending the original output.}
    \label{tab:grammar_llama}
\end{table*}
\begin{table}[!h]
    \centering
    \begin{tabularx}{\linewidth}{bsss}
        \toprule
        \textbf{Model} & \textbf{EM} & \textbf{SM} & \textbf{Corr.} \\
        \midrule
        Flan-T5-base & 0.56 & 0.65 & N/A\\
        Flan-T5-XL & 0.6 & 0.68 & 0.53\\
        Flan-T5-XXL & \textbf{0.62} & \textbf{0.72} & \textbf{0.58} \\
        Llama-2-chat-7b & 0.30 & 0.68 & \textbf{0.58} \\
        \bottomrule
    \end{tabularx}
    \caption{Exact match scores, substring match scores, and GPT correction scores for different grammar correction models.}
    \label{tab:grammar}
\end{table}
\begin{table*}[!th]
    \centering
    \begin{tabularx}{\textwidth}{sbs}
        \toprule
        \textbf{Label} & \textbf{Definition} & \textbf{Counts}\\
        \midrule
        \textbf{Unusable} & The audio is either completely silent, the speaker(s) are conversing in Mandarin, or the utterance is completely unintelligible. & 8 \\
        \textbf{Negative} & The speaker displays negative sentiments: e.g. anger, frustration, or sadness. Include instances in which the speaker displays an unwillingness to communicate. Include instances where the speaker asks for clarification, as it is an implicit display of confusion. & 39\\
        % Include instances when the speaker asks for clarifications or repetitions, as it is an implicit display of confusion.\\
        \textbf{Pauses} & The speaker makes many pauses during their utterance. These pauses make it sound like the speaker is struggling to construct the sentences. & 54\\
        \textbf{Neutral} & This includes all usable clips that are labeled neither negative nor pauses. & 200 \\
        \bottomrule
    \end{tabularx}
    \caption{Definitions for different labels in our data labeling process and their corresponding counts in our labeled audio dataset. These label definitions were presented to the annotators.}
    \label{tab:data_labels}
\end{table*}

\subsection{Grammatical Feedback}
\textbf{Grammar Correction Model: }
Following the framework in \citet{liang-etal-2023-chatback}, we train a grammar correction model to modify user utterance transcripts. We originally prompted ChatGPT for grammar correction feedback. However, responses were often hallucinated or malformed, including using the original utterance as the correction despite correctly identifying grammatical mistakes. Therefore, we train a Llama-2-7b \cite{touvron2023llama} model on ErAConD \cite{yuan-etal-2022-eracond}, which contains high-quality error-correction pairs collected from human-chatbot written dialogues. Since grammar correction is a sequence-to-sequence task, we train additional Flan-T5 models \cite{chung2022scaling} on the same data. We include more details about the training process in Appendix~\ref{app:training_detail}. 

To evaluate the models, we compute the exact match scores between model predictions and the ground truth corrections in the ErAConD test set. Llama occasionally extends its output (See Table~\ref{tab:grammar_llama}), so we include another criterion, substring match, to indicate whether the ground truth is included in the prediction. We also evaluate grammar correction quality on transcribed spoken utterances for Flan-T5-XL, XXL, and Llama. Our trained models correct 100 transcribed spoken utterances. Due to the lack of ground truth grammar corrections, we use AI feedback from GPT-4-Turbo to assess if each prediction is grammatically correct. The results of the evaluation are shown in Table~\ref{tab:grammar}.

We observe an increase from exact match to substring match across the board because the ground truth grammatical corrections do not always append periods, while most trained models do. As we transition to out-of-domain data (from written to transcription), we see a decrease in correction accuracy. However, this drop is the smallest for Llama, suggesting higher generalizability to out-of-domain data. We, therefore, choose Llama for our grammar correction model for its relatively higher robustness and smaller size than Flan-T5-XXL. 

\noindent \textbf{Grammatical Feedback Format: }We would like to present grammar model corrections to the students. Upon considering our design survey results, we choose conversational recasts \cite{lyster2013oral}. This involves reformulating student utterances, often including confirmation checks (e.g. "Did you mean [corrected sentence]?"). We implement the recast by pre-pending the corrected sentence with a random confirmation check phrase (e.g. "I think you meant"). When the corrected sentence is longer than 20 words, we instead identify a dependency parse constituent containing the error to avoid repeating the entire sentence when possible. Since the sentences are sentence-tokenized before being corrected, we ignore Llama corrections longer than one sentence. This addresses the previous Llama extension issue.

In addition to a conversational recast, we want to explain how the student's utterance is incorrect. We utilize the conversational grammar correction feedback templates proposed in \citet{liang-etal-2023-chatback} and append the templated feedback to the utterance. See examples of our grammatical feedback in Appendix~\ref{app:example_conversation}.

\subsection{Negative Emotion and Pause Detection}

\textbf{Data Labeling:} Since no accented speech emotion classification dataset exists, we labeled our audio clips to create evaluation data for our pipeline. We used four labels: Negative, Pauses, Neutral, and Unusable. Two Mandarin native speakers with high English proficiency annotated approximately 10\% of the data with a Kappa of 0.893. We only include audio clips whose labels both annotators agreed upon. Our audio dataset's data distribution and label definitions are in Table~\ref{tab:data_labels}. The label definitions were presented to the annotators as the annotation scheme. The annotators also labeled clips featuring both negative affect and pauses as ``Negative'' to promote better label balance, since students rarely display negative emotions in our data.

\noindent \textbf{Negative Emotion Detection:} Because of the shortage of emotion-labeled accented speech data, we could not train new audio classification models for our specific task. Instead, we manipulate a popular out-of-the-box speech emotion classification model\footnote{https://huggingface.co/ehcalabres/wav2vec2-lg-xlsr-en-speech-emotion-recognition}. We test different configurations and settings for this model on the small emotion-labeled dataset from the previous segment. Specifically, given the output probabilities for different emotions, we first combine a subset of them to form our estimated probability of negative affect. Thresholding is then applied to provide the final prediction. Our results indicate that the optimal configuration is the predicted probability for only ``anger'' and a threshold of 0.4. Using this information, we reach a weighted F1 score of 0.78 on our Negative and Neutral audio clips. See Appendix~\ref{app:detail_emo} for details.

The current speech emotion recognition models do not perform well on our task, as speculated.
Anecdotally, when directly running classification on our audio clips using the model, many clips classified by us as ``Neutral'' are often classified as sad or disgusted.

\begin{table}[!t]
    \centering
    \begin{tabularx}{\linewidth}{XXms}
        \toprule
        \textbf{Metric} & \textbf{Clip Label} & \textbf{Mean} & \textbf{Diff}\\
        \midrule
        Ratio & Pauses &  0.41 & \\
         & Neutral & 0.32 & 0.09\\
        Pause Rate & Pauses & 0.60 & \\
        & Neutral & 0.55 & 0.05 \\
        Pause Length & Pauses & 0.68 & \\
         & Neutral & 0.49 & \textbf{0.19}\\
        \bottomrule
    \end{tabularx}
    \caption{The three metrics for the clips labeled as ``Pauses'' and ``Neutral'' in our audio data. We include the average values for the metrics, as well as the differences between the different clip labels.}
    \label{tab:pauses}
\end{table}

\noindent \textbf{Pause Detection:} 
As established, prolonged pauses indicate the presence of foreign language anxiety and should be considered as a cue in our framework. We aim to develop automated metrics that identify user utterances with these pauses.

We devise three metrics for determining whether an audio clip fulfills the criteria for ``Pauses'':
\begin{enumerate}
    \item \textbf{Silence Ratio:} The quotient of the total amount of silence in a clip and the clip length.
    \item \textbf{Pause Rate:} The result of dividing the number of pauses by audio length.
    \item \textbf{Average Pause Length:} The average length of pauses.
\end{enumerate}

For computing these metrics, we equip our system with Silero-VAD \cite{Silero-VAD}, a fast and enterprise-grade voice activity detection package. Silero-VAD identifies and locates speech segments, and it allows speech extraction from the original audio such that the resulting clip is speech-only. We can therefore compute the total lengths of silence and pauses in an audio, as well as the number of pauses in an audio clip. 

Other features, such as pause location, can also be used to indicate the level of anxiety. We leave the exploration of these features to future work.

To compare the ability of these metrics to differentiate between ``Neutral'' and ``Pauses'' clips, we calculate the values of these three metrics on these clips. We further measure the differences between the metric values for the two categories (Table~\ref{tab:pauses}). The ``Average Pause Length'' metric yields the highest difference, which suggests it effectively separates ``Neutral'' and ``Pauses'' clips. 
In addition, we experiment with various thresholds for differentiating the two types of audio using ``Average Pause Length'' (See Appendix~\ref{app:pause_detection}) and select a threshold of 0.5. 

\subsection{Empathetic Response Generation}
\textbf{Data Construction:} Given the ASR transcripts of user utterances in a conversation, we added all instances of three consecutive utterances to our data (i.e., utterances 1+2+3, 2+3+4, etc). This created 2014 segments for optimizing our ChatGPT prompts. Due to cost constraints, we only used 625 conversation segments for prompt optimization: 125 for optimization, 200 for evaluation and iteration, and 300 for held-out testing.

\noindent \textbf{Implementation:} Our desiderata for the empathetic response generation module include the following: 1. Tailored to the user; 2. Empathetic and encouraging; 3. Including actionable feedback or specific examples the user can learn from. Because there are no sufficiently large datasets that precisely fulfill these requirements, we rely on prompting ChatGPT to generate such responses. Unfortunately, large language models are sensitive to how they are prompted. Simple trial-and-error did not achieve consistently satisfactory responses in our preliminary experiments (\textsc{Zeroshot} stage). 

We employed the DSPy framework \cite{khattab2023dspy} to optimize for prompts while satisfying our desiderata (\textsc{Optimized} stage). We first tasked GPT-4 to check whether each requirement is satisfied in a given response (e.g. is the utterance empathetic and encouraging). This is a form of AI feedback \cite{bai2022constitutional}. GPT-4 appears successful in this text annotation task, consistent with results established in \citet{gilardi2023chatgpt}. Using the AI feedback as our \textit{metrics}, we aimed to optimize our prompts to maximize the metrics. DSPy supplies the BayesianSignatureOptimizer, which references simple descriptions of our desiderata to suggest sample instructions and few-shot examples. Using this Bayesian-model-powered optimization process, we improved the metrics on a held-out test set from 68.3 (at the \textsc{Zeroshot} stage) to 89.8. We discuss whether the improvement aligns with human intuition in Section~\ref{sec:emp_gen_eval}.

We observe that the outputs of our optimized prompt are often formal, while most of our design survey participants prefer colloquial feedback. To address this, we insert a final rewrite call to rewrite the optimized prompt output to a more colloquial version (\textsc{Rewrite} stage). GPT-4 evaluates this stage's outputs as 88.7.

During inference time, when we detect that the user requires empathetic feedback, the user's three most recent utterances are concatenated and fed into ChatGPT with the optimized prompt. The output undergoes the \textsc{Rewrite} stage to produce the final output. 
All ChatGPT prompts and GPT-4 feedback prompts used for this module are included in Appendix~\ref{sec:emp_feedback_prompts}. See Appendix~\ref{app:examples} for examples of outputs at different stages.

\subsection{Connecting Feedback and Conversation}
\textbf{User Query Response:} Our feedback modules are currently intended for single-turn feedback (i.e. the bot provides the feedback without anticipating that the user will ask clarification questions), but in preliminary user studies, we noticed that users do inquire about the feedback. Therefore, we handle this case by constructing a ChatGPT call with the immediate conversation context and asking for a response to the user's query. We classify a user response to feedback as a relevant query with a rule-based approach. We use this rule-based approach instead of forwarding all post-feedback user queries to ChatGPT because prior users would ask about the bot's creator and training data, resulting in unintended behavior (e.g. the bot claiming it is created by Google or OpenAI engineers).

\noindent \textbf{Transition:} We employ templates for a smooth transition between feedback and the original conversation.  Before entering the feedback stage, we cache the original bot response to return to the conversation afterward. More details about templates and ChatGPT prompts are provided in Appendix~\ref{app:detail_connect}.

\subsection{Conversation}
Unlike the other modules that only need to be activated sporadically, the conversation module is invoked for almost every turn. This poses additional needs for inference speed and costs, which motivates using a locally stored model.

We selected a Vicuna model fine-tuned for curriculum-driven conversations \cite{li2023curriculum}. The model allows for customization of topics, chatbot personas, and vocabulary to incorporate into the conversation. \citet{li2023curriculum} noticed that brevity instructions are sometimes ignored by ChatGPT. This further makes ChatGPT not ideal for our spoken conversation use-case, as run-on utterances may be difficult to comprehend in a speech setting. 
Users found the Vicuna model more helpful for developing conversational skills, providing natural and realistic utterances, and aligning with users' English proficiency levels. 

The topic of ``Name a movie that has had an enduring impact on you'' was chosen for relatability. We randomly selected a vocabulary and one of the female personas to match the TTS voice. Bot feedback and user responses to feedback are not included in the conversation history when prompting the Vicuna model to keep the components modular and prevent out-of-distribution behavior. 

\begin{table*}[!t]
    \centering
    \begin{tabularx}{\linewidth}{XXXXXXXXX}
        \toprule
         & \textbf{Quality} & \textbf{Conf.} & \textbf{Useful} & \textbf{Enc.} & \textbf{Listen} & \textbf{Care} & \textbf{Praise} & \textbf{PAS}\\
        \midrule
        \textbf{Average} & 3.75&3.33 & 3.83& 3.16& 3.58& 3.08&3.25 & 3.27\\
        \textbf{Std} &1.05 &1.07 & 1.19&1.64 &1.16 &1.24 & 1.60& 1.16 \\
        \bottomrule
    \end{tabularx}
    \caption{Post-survey results. "Conf." stands for confidence, "Enc." stands for encourage, and "PAS" stands for perceived affective support.}
    \label{tab:post_survey}
\end{table*}

\section{User Study}

Fourteen native Mandarin speakers were recruited from social media and the authors' connections. Each participant conversed with the chatbot for at least 10 turns (a turn is one round of exchange between the chatbot and the user). A pre-survey for participant English proficiency and a post-survey for user experience were administered. In the pre-survey, we obtain an approximate assessment of the participants' English proficiency including their standardized test scores, self-reported proficiency, and the frequency at which they speak English daily. After the participants interacted with our system, they were presented with a post-survey which includes a modified version of the teacher affective support scale (the last four items below) \cite{sakiz2007does} adapted for our context and general evaluations of conversation quality. 

Our Likert-scale post-survey includes: 

\noindent \textbf{Quality:} How was the conversation quality?

\noindent \textbf{Confidence:} Do you feel that you are more confident after conversing with the chatbot?

\noindent \textbf{Useful:} Do you think the chatbot's grammar feedback is useful?

\noindent \textbf{Encourage:} The chatbot encourages me when I am having difficulties in the conversation.

\noindent \textbf{Listen:} The chatbot listens to me when I have something to say.

\noindent \textbf{Care:} My opinion matters to the chatbot.

\noindent \textbf{Praise:} The chatbot recognizes and appreciates when I am good at something.

Details for the surveys can be found in Appendix~\ref{sec:pre_post}. Example conversations between the participants and the bot can be found in Appendix~\ref{app:example_conversation}.

\begin{table}[!t]
    \centering
    \begin{tabularx}{\linewidth}{XXXX}
        \toprule
        \textbf{Stage} & \textbf{vs. \textsc{Zeroshot}} & \textbf{vs. \textsc{Optimized}} & \textbf{vs. \textsc{Rewrite}}\\
        \midrule
        \textsc{Zeroshot} & - & 0.52 & 0.45\\
        \textsc{Optimized} & 0.47 & - & 0.45 \\
        \textsc{Rewrite} & 0.54 & 0.54 & - \\
        \bottomrule
    \end{tabularx}
    \caption{Win rates between each pair of empathetic feedback generation stages. }
    \label{tab:ranks}
\end{table}

\section{Results and Discussion}

\subsection{Empathetic Generation Evaluation}
\label{sec:emp_gen_eval}

We asked each participant to rank the different stages of empathetic feedback (\textsc{Zeroshot}, \textsc{Optimized}, \textsc{Rewrite}). Participants ranked responses generated in these three stages on the same segment for 30 randomly selected segments. 
At least 3 participants ranked each triple. We also asked the participants how they would improve the utterance they ranked at \#1 for each conversation segment. 

In Table~\ref{tab:ranks}, \textsc{Rewrite} wins more often against both \textsc{Zeroshot} and \textsc{Optimized}, suggesting that the \textsc{Rewrite} improves \textsc{Optimized} stage outputs. \textsc{Optimized} outputs are often not preferred due to their formality and length. Since \textsc{Rewrite} rephrases \textsc{Optimized} outputs without modifying core content, it appears that the participants are ranking the content from \textsc{Optimized} relatively higher than the content from \textsc{Zeroshot}. Another result is that \textsc{Zeroshot} is often ranked as \#1 or \#3, illustrating that \textsc{Zeroshot} outputs are less consistent in quality. Despite being scored higher by GPT-4, \textsc{Optimized} does not significantly outperform \textsc{Zeroshot}. This could be due to DSPy optimization focusing on fulfilling metrics without considering human preferences, or due to raters having various standards.

As for improving the feedback, participants reported that the best responses are still too verbose (one wrote "the shorter the better") and requested better feedback examples. They mentioned that generic praises can sound disingenuous, detrimental to the intention to encourage. Some suggested that praise may not be necessary for every piece of feedback, especially when participants receive multiple feedback during a conversation. One future direction would be to develop more context-aware mechanisms for more naturalistic and long-term empathetic feedback.

\subsection{Conversation Statistics}

Two participants did not receive empathetic feedback and were excluded from analyses. For the other twelve participants, each conversed for an average of 14.5 turns and received 1.9 grammatical feedbacks and 1.3 empathetic feedbacks.

\subsection{Survey Results}
On average, our participants have approximately 14.25 years of experience learning English. They all rated themselves above three out of five for self-reported English proficiency (higher is more proficient) with an average of 3.92. The participants who disclosed their IELTS and TOEFL scores had 7.3 and 109.3 averages respectively. For the question on English usage frequency, the average was 3.41 (one being for English only, five being for Mandarin only). Our participants have intermediate English proficiency but do not speak English frequently. 

The post-survey results are shown in Table~\ref{tab:post_survey}. In addition to the survey items, we include PAS as an aggregate metric by averaging the four adapted PAS survey items. The participants often consider the conversation quality to be high. They reported gaining moderate confidence after the conversation, and consider the bot's feedback useful. As for the survey items involving PAS, the results contain higher variance. While users believe that the bot appears to listen to them fairly attentively (potentially as an effect of the grammatical feedback), they are more ambivalent about whether the bot encourages them or praises them appropriately. We suspect that the reason for lower ``Encouragement'' ratings stems from our imperfect detection mechanism; empathetic feedback might have been given when the user was not exactly struggling. The participants also could not have struggled at all during the conversation. A potential reason for the high variance in ``Praise'' ratings is the disingenuous-sounding encouragement mentioned in Section~\ref{sec:emp_gen_eval}. Additionally, user motivation for using our system can affect their self-reported results. Participants who only intend to test the system rather than improve their English might rate it poorly.

\subsection{Dialogue Inspection and User Feedback}

We inspect conversations with low PAS to identify failure modes of our system. The conversation with the lowest PAS includes both technical issues in the system (the user was baffled by the frequent interruptions in the system) and the chatbot forgetting the conversation history due to the limited context length of our model. Another conversation features significant ASR errors and the error propagation led to nonsensical grammatical feedback which confused the user. Due to current limitations in user query processing after bot feedback, some user queries were occasionally ignored, but the presence of these does not dictate low PAS.

We requested feedback from our participants. They praised the ASR accuracy and feedback quality, mentioning that they feel encouraged after receiving feedback. Some users stated that the goal of spoken English is to keep the conversation going, and therefore only egregious grammar errors should be corrected. Others would solicit grammar feedback from the system and exhibit dismay when it did not recognize their errors. One user mentioned that they would stammer and have disfluencies that would be recognized as grammatical errors. Some users disliked the stiffness of the feedback formats as they felt the conversation flows were interrupted. A subset of responses are presented in Appendix~\ref{app:user_feedback}. 

These observations highlight limitations in our current system. To improve user experience, we will develop more seamless feedback mechanisms and robust user query classification. Additionally, we aim to create better grammar models suited for transcribed utterances and resilient to disfluencies and fillers. Additional goals include detecting technical difficulties so the chatbot can apologize for any interruption, as well as conversation summarizers to inform our model of previous discussions. 

\section{Conclusion and Future Work}

In this work, we propose the negative emotion detection task in the context of English learning to capture learner frustration and anxiety. We also introduce the first English-teaching chatbot that provides adaptive, empathetic feedback to students using our negative affect detection pipeline. Initial trials with end users demonstrate the potential of our system. For future work, we intend to scale up our user evaluations and verify our hypothesis that our system can effectively improve student L2 grit.

For future work, apart from addressing participant feedback, we intend to expand our experiments to include more thorough comparisons between the different experimental conditions to establish more robust results. Specifically, we want to determine whether our adaptive empathetic feedback improves L2 grit more than no empathetic feedback or fixed feedback upon multiple chatbot interactions. Another interesting topic to examine more closely would be whether humans behave and react similarly when conversing with chatbots and real-life English teachers. We intend to include participants from an ESL course in our next study.

\section{Limitations}
Our current system serves as a proof-of-concept for a chatbot system capable of adaptive empathetic feedback, and it is by no means perfect. While our modular design allows for more rigorous control for future experiments, there can easily be error propagation between modules, and none of the modules are completely error-proof, as we have illustrated in our paper. To begin with, our speech emotion recognition pipeline does not successfully capture all instances of negative affect in our labeled data. The Llama model used for grammar correction still cannot correct all instances in the ErAConD test set. Our user query detection mechanism can miss relevant queries. All of these should be improved in future iterations of the system.

The current user study results are preliminary and do not offer sufficient statistical strength for solid conclusions. In future, we will aim for larger user studies by recruiting broadly on social media and at our institution. 

Our data is currently labeled only by two labelers, which renders our labels less valid. We will aim to include more labelers to improve the validity of our emotion-labeled data. 

\section{Ethical Considerations}
Any applications interfacing with humans, especially students, need to consider accidental psychological harm done to the students as a result of generations. To address this, we performed rigorous testing prior to our user study.

There is potentially self-identifying information present in our audio data. We will filter out self-identifying information before releasing the data to protect user identity.

% Bibliography entries for the entire Anthology, followed by custom entries
%\bibliography{anthology,custom}
% Custom bibliography entries only
\bibliography{anthology,custom}

\appendix

\input{appendix}

\end{document}

%% file: appendix.tex
\section{Chatbot Design Discussion}
\label{app:discussion}

A total of 12 Mandarin native speakers were recruited through the authors' personal connections to provide feedback on our chatbot design. We provided a Google Form for them to complete. We later released a version on social media that gained many more responses, but since we do not utilize the results from that survey directly in this work, we reserve the discussion and incorporation of these results for future work.

We translated a subset of relevant questions and response options from Mandarin. We have abridged preambles in the questionnaire for brevity. The questions and responses of our survey are as follows (the percentage in the parentheses corresponds to the percentage of participants who selected that option):
\begin{enumerate}
    \item \textit{How would you like an English teacher to give you feedback?}
    \begin{itemize}
        \item Formal (25\%)
        \item \textbf{Colloquial (75\%)}
    \end{itemize}
    \item \textit{How long should the feedback be?}
    \begin{itemize}
        \item 1 - 2 sentences (8.3\%)
        \item 2 - 3 sentences (41.7\%)
        \item \textbf{3 - 4 sentences (50\%)}
        \item 4+ sentences (0\%)
    \end{itemize}
    \item If you have made a grammatical error, what specific attributes should a teacher's feedback for you have? Please select all that are applicable.
    \begin{itemize}
        \item Correct your mistakes directly (58.3\%)
        \item Correct your mistakes interactively through Q \& A (33.3\%)
        \item \textbf{Give you examples so that you can learn from the example and avoid making the same mistakes again (75\%)}
    \end{itemize}
    \item What is your ideal form of encouraging and empathetic feedback? Please select all that apply.
    \begin{itemize}
        \item Give you encouragement (e.g. "You are doing great!", "I am proud of you!") (58.3\%)
        \item \textbf{Tell you what you are good at in terms of your spoken English (75\%)}
        \item Tell you what you can improve in your spoken English (58.3\%)
        \item \textbf{Tell you how you can improve through examples (75\%)}
        \item Provide you with plausible and actionable advice for improving your English (41.7\%)
    \end{itemize}
\end{enumerate}

\section{Experiments for Pause Detection}
\label{app:pause_detection}

After computing the pause length metric values for all audio clips labeled as either ``neutral'' or ``pauses'', we compared the effect of varying pause detection mechanisms and thresholds. We computed the classification accuracy values when using different pause detection metrics and different threshold values. We aim to obtain the highest possible classification accuracy values for our metric-threshold combination. 

We present the results for varying threshold values for Silence Ratio, Pause Rate, and Average Pause Length in Tables \ref{tab:silence_ratio}, \ref{tab:pause_rate}, and \ref{tab:average_pause_length}, respectively.

\begin{table}[!t]
    \centering
    \begin{tabularx}{\linewidth}{XXX}
        \toprule
        \textbf{Threshold} & \textbf{Neutral\%} & \textbf{Pauses\%}\\
        \midrule
        0.1 & 100.0 & 3.5 \\
        0.2 & 98.1 & 22.5 \\
        0.3 & 72.2 & 53.0 \\
        0.4 & 44.4 & 72.5\\
        0.5 & 26.0 & 85.5 \\
        0.6 & 7.4 & 92.0 \\
        0.7 & 3.7 & 97.0\\
        \bottomrule
    \end{tabularx}
    \caption{Classification accuracy for Neutral and Pauses audio clips using the Silence Ratio pause detection metric.}
    \label{tab:silence_ratio}
\end{table}

\begin{table}[!t]
    \centering
    \begin{tabularx}{\linewidth}{XXX}
        \toprule
        \textbf{Threshold} & \textbf{Neutral\%} & \textbf{Pauses\%}\\
        \midrule
        0.1 & 100.0 & 0.0\\
        0.2 & 98.1& 1.5 \\
        0.3 & 96.3 & 9.5 \\
        0.4 & 88.9 & 18.5\\
        0.5 & 74.1 & 39.5\\
        0.6 & 50.0& 61.0\\
        0.7 & 29.6& 85.0\\
        \bottomrule
    \end{tabularx}
    \caption{Classification accuracy for Neutral and Pauses audio clips using the Pause Rate pause detection metric.}
    \label{tab:pause_rate}
\end{table}

\begin{table}[!t]
    \centering
    \begin{tabularx}{\linewidth}{XXX}
        \toprule
        \textbf{Threshold} & \textbf{Neutral\%} & \textbf{Pauses\%}\\
        \midrule
        0.1 & 98.1 & 21.5\\
        0.2 & 98.1 & 24 \\
        0.3 & 96.3 & 34.5 \\
        0.4 & 88.9 & 54.4 \\
        0.5 & \textbf{70.4} & \textbf{73.0} \\
        0.6 & 61.1 & 81.5 \\
        0.7 & 38.9 & 89.5\\
        \bottomrule
    \end{tabularx}
    \caption{Classification accuracy for Neutral and Pauses audio clips using the Average Pause Length pause detection metric.}
    \label{tab:average_pause_length}
\end{table}

\section{Experiments for Negative Emotion Detection}
\label{app:detail_emo}

The wav2vec model we have selected outputs probabilities for the following emotions given a speech segment: angry, calm, disgust, fearful, happy, neutral, sad, and surprised. This model is a fine-tuned version of \citet{grosman2021xlsr53-large-english}, which is a widely popular automatic speech recognition model. The model is then fine-tuned on the RAVDESS dataset \cite{livingstone2018ryerson} for the speech emotion recognition downstream task. 

We explored the following methods for combining the output probabilities from the model to produce the negative affect estimate. Here, we include anger for each of our methods because frustration and anger can manifest themselves in a similar manner in speech. 

\begin{enumerate}
    \item Anger + Disgust + Fearful + Sad (ADFS) - 0
    \item Anger + Disgust + Fearful (ADF) - 1
    \item Anger + Disgust (AD) - 4
    \item Anger + Fearful (AF) - 5
    \item Disgust + Fearful (DF) - 3
    \item Anger (A) - 2
\end{enumerate}

For the values of the threshold, we experimented with 0.1 through 0.9 with an increment of 0.1.

We evaluated the different setups on all audio clips labeled as Neutral or Negative in our labeled data subset. The weighted F1 score was computed to account for class imbalance. We included the best F1 scores achievable by each setup, along with their corresponding thresholds for obtaining the best F1 scores, in Table~\ref{tab:emo_exp}.

\begin{table}[!t]
    \centering
    \begin{tabularx}{\linewidth}{XXX}
        \toprule
        \textbf{Setup} & \textbf{Threshold} & \textbf{Best F1}\\
        \midrule
        ADFS & 0.9 & 0.57\\
        ADF & 0.8 & 0.76 \\
        AD &  0.8 & 0.76\\
        AF & 0.4 & 0.76 \\
        DF & 0.8 & 0.76 \\
        A & 0.4 & \textbf{0.78} \\
        \bottomrule
    \end{tabularx}
    \caption{The best achievable weighted F1 scores and their corresponding threshold values for each of the speech recognition model output aggregation methods.}
    \label{tab:emo_exp}
\end{table}

\section{Training Details for Llama and Flan-T5 Models} \label{app:training_detail}
All models were trained on a single 40 GB GPU. Models were trained for 10 epochs and the best models were selected using validation set loss. Flan-T5-base was trained without any parameter-efficient fine-tuning, but all other models were trained using PEFT and Lora. We will release model training and inference code if accepted.

\section{Details on the Connecting Feedback and Conversation Component}
\label{app:detail_connect}
\subsection{Identifying Query}
We utilized a simple rule-based approach to determine whether an utterance provided by a user after bot feedback is a question about the feedback or the English learning process. Namely, we (1) checked that a "?" is in the transcribed utterance; (2) checked whether one of the words in this list belongs in the utterance: "grammar", "grammatical", "vocab", "English", "mistake", "example", "sentence". If both conditions were fulfilled, we then interfaced with ChatGPT to respond to user queries.

\subsection{Responding to Query}

Since we anticipate that the user will only be responding to the bot feedback, we would not need to include too much context in our ChatGPT call. We supplied the following prompt to ChatGPT to obtain a response to give to the user.

\begin{lstlisting}
    f"""Based on the following conversation history:\n\n{convo}, answer the user's following query: "{user_query}" Answer in a spoken utterance. Provide specific feedback, but be succinct."""
\end{lstlisting}

\subsection{Transitioning}

If the user does not respond to the bot feedback with a query, or when the bot has finished responding to the user query, the system would then return to the original conversation flow. This transition was obtained by prefixing the cached original bot response with one of the randomly selected phrases. There are two general scenarios here:
\begin{enumerate}
    \item The user expresses gratitude by including "thank" in their utterance.
    \item The user does not explicitly express gratitude.
\end{enumerate}

The code for constructing the prefix to prepend to the original bot response is as follows:
\begin{lstlisting}
        if "thank" in text.lower():
            prefix = random.choice(["Of course!", "No problem at all.", "Yeah, no problem!", "No problem!"]) + " " + random.choice(["Back to the conversation.", "Back to our convo.", "Let's go back to chatting.", "Now we circle back."])
        else:
            prefix = random.choice(
                ["Sounds great.", "Alright, let's continue our conversation.", "Great, let's get back to it!",
                 "Okay let's go back to our conversation.", "Now back to our conversation.", "Okay!",
                 "Lets' go back to our chat.", "Let's keep chatting."])
\end{lstlisting}

\section{User Feedback}
\label{app:user_feedback}
We asked participants to share their experience with the chatbot, and received the following responses. All responses have been paraphrased and translated from Mandarin:

\textit{\textbf{User A:} I like the chatbot and I look forward to the finished product... When the chatbot corrects my grammar, the way it corrects me is very stiff and clunky, and it can interfere with the immersive experience... Additionally, when I use filler words or discourse markers or perform self-repair in the conversation, the bot would catch these as grammatical errors even though they should not be treated as such.}

\textit{\textbf{User B:} Your chatbot has high ASR accuracy, and the quality of the conversation is quite good! However, it may be challenging to provide English learners with sufficient help, because I asked for more specific feedback after receiving a feedback and it went back to talking about movies.}

\textit{\textbf{User C:} One thing I really like about the chatbot is that the feedback is very timely! When using ChatGPT I don't always get very prompt feedback. We have a foreigner English teacher coming next week and your bot is definitely a life-saver.}

\textit{\textbf{User D:} I don't think some trivial grammatical errors, like mixing is and was when discussing movie plots, should be corrected. I also asked for the bot for word choice improvements but it didn't give me any.}

\textit{\textbf{User E:} The conversation has high quality, and although the feedback can lag behind one or two sentences, the feedback is always correct and helpful. The bot clearly understands what you are trying to convey and can continue the conversation despite grammar errors. Sometimes it uses sentences that can be too complex for beginners, but providing Mandarin translations for bot utterances can be helpful.}

\section{Example Conversations}
\label{app:example_conversation}

The following are conversation excerpts between our bot and different users. Here, {\color{blue} blue} indicates empathetic feedback, {\color{orange} orange} indicates a transition between feedback stage and the conversation, and {\color{red} red} indicates a grammatical feedback. 

We first examine a successful conversation example. Note that the empathetic feedback is activated when the user experiences hesitation:

\noindent \textbf{User:} Actually, I hardly ever watch movies, so could I describe opera?

\noindent \textbf{Bot:} Sure! What's the name of the opera that you'd like to describe?

\noindent \textbf{User:} Okay, that's Turandot, which describes a love story between a Chinese princess and a foreign prince.

\noindent \textbf{Bot:} {\color{red}I believe you wanted to say "Okay, that's Turandot, which is a story about a love between a Chinese princess and a foreign prince.". In this sentence you used the verb "describes", but it doesn't sound natural. I'd recommend using the word "is a story about".  Does that sound good?}

\noindent \textbf{User:} you

\noindent \textbf{Bot:} {\color{orange}Okay let's go back to our conversation.} Interesting! I can't say that I'm familiar with it. Could you tell me more about the story?

\noindent \textbf{User:} This Chinese princess whose grandma is... Wait a minute.

\noindent \textbf{Bot:} {\color{blue} You've got a good grasp of the topic and can explain yourself clearly, which is awesome! Just tweak your grammar and sentence structure for a more natural flow. For instance, instead of saying "I hardly ever watch movies, so could I describe opera?", try "Since I rarely watch movies, can I talk about opera instead?" Keep at it, and you'll see improvement in no time!  Does that sound alright to you?}

\noindent \textbf{User:} Yes, and now I will go to talk about the context of the opera.

\noindent \textbf{Bot:} {\color{orange} Okay!} Yes, go on. What's the story about?

\noindent \textbf{User:} This story is about the Chinese princess, Truong Du, whose grandma was bullied by the foreigners. So Truong Du set a rule to the man who want to marry him that he must answer three questions and then he can marry her or dad. The cover of the, answered the three questions and finally married with Torandu.

\noindent \textbf{Bot:} {\color{red} Perhaps what you meant to say was "who wants " and not want In this sentence you made a mistake on the verb "want". The correct verb form here is "wants". Remember to make your verbs agree with their subjects.  How does that sound? }

\noindent \textbf{User:} That sounds great, okay I understand.

\noindent \textbf{Bot:} {\color{orange} Alright, let's continue our conversation.} That's certainly a unique story! What did you like most about the opera?

\noindent\rule{6cm}{0.4pt}

Here we see a failure of recognizing user query due to the rule-based nature of our query recognition mechanism. 

\noindent \textbf{Bot:} I'm doing alright, thanks. Have you watched any good movies lately?

\noindent \textbf{User:} No, not really. I don't watch any movie.

\noindent \textbf{Bot:} Really? You don't watch any movies? Not even the classics?

\noindent \textbf{User:} then it won't be lately. I do watch classics. Which one you want to talk about?

\noindent \textbf{Bot:} Well, I was thinking about the movie "Citizen Kane". What do you think about it?

\noindent \textbf{User:} That is a very very classical one. I had it some time ago I couldn't really remember the plot could you rephrase it for me?

\noindent \textbf{Bot:} Sure, it's about a journalist who becomes a powerful newspaper publisher and his journey to find the meaning of "rosebud" which holds the key to his soul.

\noindent \textbf{User:} Yeah, I sort of like that idea. You know, start being a nobody and then with one's effort you became a big shot in the industry.

\noindent \textbf{Bot:} Exactly, it's a great story of how one's perseverance and hard work can lead to success. I also love the way the director used different narrative techniques to tell the story.

\noindent \textbf{User:} I couldn't remember exactly. Could you give me an example?

\noindent \textbf{Bot:} Sure, one of the techniques used is the use of different reporters telling the story of Kane's life. Each one provides a different perspective on who he was as a person and the events that shaped his life.

\noindent \textbf{User:} Yeah I would say that was quite innovative at that time when the movie was first released. Anyway what else you want to talk about?

\noindent \textbf{Bot:} {\color{blue} You're doing great with understanding and expressing the topic clearly! Just brush up on grammar and sentence structure, like saying "I can't remember exactly." Keep practicing to keep getting better!}

\noindent \textbf{User:} All right, anything else for me?

\noindent \textbf{Bot:} {\color{orange} Lets' go back to our chat.} Well, I was also thinking about "The Godfather". Have you seen it?

\noindent \textbf{User:} Oh yes, I really love the music. The Godfather theme in both 1 and 2. I didn't really watch Godfather, the third part.

\noindent \textbf{Bot:} {\color{red} I believe you wanted to say "I didn't really watch The Godfather, the third part.". You seem to be missing a determiner in this sentence. You should probably add "The" to make the sentence sound more natural.  Does that sound alright to you?}

\section{Pre Survey and Post Survey}
\label{sec:pre_post}
The surveys have been translated from Mandarin.
\subsection{Pre-Survey}
\begin{enumerate}
    \item How many years have you been learning English?
    \item How would you rate your English proficiency (1 star for beginner, 5 stars for proficient)?
    \item What is your IELTS score?
    \item What is your TOEFL score?
    \item How much English do you speak every day (1 star for English only, 5 stars for Mandarin only)?
\end{enumerate}

\subsection{Post-Survey} 
\begin{enumerate}
    \item How was the quality of your conversation on a scale of 1 to 5?
    \item Do you feel that you are more confident after conversing with the chatbot? Rate on a scale of 1 to 5.
    \item Do you think the chatbot's grammar feedback is useful? Rate on a scale of 1 to 5.
    \item The chatbot encourages me when I am having difficulties in the conversation (1-star for "Not like me at all", 5-stars for "Very much like me").
    \item The chatbot listens to me when I have something to say (1-star for "Not like me at all", 5-stars for "Very much like me").
    \item My opinion matters to the chatbot (1-star for "Not like me at all", 5-stars for "Very much like me").
    \item The chatbot recognizes and appreciates when I am good at something (1-star for "Not like me at all", 5-stars for "Very much like me").
\end{enumerate}

\section{Empathetic Feedback Generation ChatGPT Prompts}
\label{sec:emp_feedback_prompts}
\subsection{Zero-shot Prompt}

\begin{lstlisting}

A student is learning English. You are assessing a spoken utterance. In at most two sentences, summarize (1) their specific strengths in English skills and (2) things they can work on to improve. Address the student in the second person. Include specific examples that the student can learn from. Be colloquial, as if in spoken conversation.

---

Follow the following format.

Convo: \$\{convo\}
Reasoning: Let's think step by step in order to \${produce the output}. We ...
Output: Treat this as a spoken conversation, so be succinct, colloquial, and empathetic.

---

Convo: - Yeah, I always talk to my parents about the student. - Yeah, so the person that I met, you know, that I've met, that I want to study with is one of my classmates, Tom, and how I met this, met Tom is when I was in high school and he's in my next classroom and he is one of the best students in the other class and I really admire his intelligence and I learn a lot from him and I've known him since, you know, 2016 and he is one of the best students in the other class so that's why I want to study with him.
Reasoning: Let's think step by step in order to assess your spoken utterance. We can see that you have a strong grasp of English vocabulary and grammar, as evidenced by your ability to express complex ideas and use a variety of sentence structures. However, you may want to work on your pronunciation and intonation, as some of your words were not clear and your speech lacked natural rhythm. For example, you said "met" instead of "meet" and "classroom" instead of "classmate." Practicing with a native speaker or using online resources can help you improve in this area. Keep up the good work!
Output:
\end{lstlisting}

\subsection{Optimized Prompt}
\input{dspy_prompt}

\subsection{Rewrite Prompts}
There are two stages to the rewrite process. In the first stage, we employed the following prompt:

\begin{lstlisting}
Shorten and rewrite this utterance to sound simple, natural, and engaging; remove any assessment of speech including pronunciation and intonation:\n\n{empathetic_output}
\end{lstlisting}

Here, \texttt{empathetic\_output} refers to the output from the \textsc{Optimized} stage. We then continued appending the following message to the current session with ChatGPT:
\begin{lstlisting}
Make your response different and casual, and shorten to 3 - 4 sentences
\end{lstlisting}

The output from ChatGPT after this additional prompting was used as the final output.

\section{Example Outputs at Different Stages}
\label{app:examples}
The following conversation segments are selected at random and may contain ASR errors.

\input{convo_examples}

%% file: dspy_prompt.tex
\begin{lstlisting}
Proposed Instruction: You're playing the role of an encouraging English tutor for a student who is actively learning and practicing their English through conversation. Your task is to listen attentively to their spoken utterances and provide constructive feedback. In your response, kindly highlight (1) one specific strength they showed or an aspect they did well in during the conversation, complimenting their effort or skill in English, and (2) offer one focused suggestion on how they can improve further, making it actionable and clear. Use colloquial language to maintain the conversational tone, directly addressing the student with "you", and where possible, reference specific examples from their speech to illustrate your points. Your feedback should feel like a supportive nudge towards their language learning journey, keeping it concise and personalized.

---

Follow the following format.

Convo: ${convo}
Reasoning: Let's think step by step in order to ${produce the output}. We ...
Feedback: Treat this as a spoken conversation, so be succinct, colloquial, and empathetic.

---

Convo: - Sorry, I have not get some information about the personal medical data sharing. Well, I think it's maybe something related to digital. Like, it's different from the traditional medical system. Like, all medical systems transfer from the traditional one into the digital one. And maybe the digital systems will help the hospital to organize the patients and to regulate some patient's information with high efficiency and better orders. So it may saves a lot of time for some... - Yeah, maybe like if you go to one hospital and if the doctor keeps a capital record on his on the paper then you left this hospital go to another hospital the new doctor have no idea about your situation if all information of patients can put on the website or in other words those are online information that whether we change the hospital the doctor can get access to our information and we can show our situation to to those doctors yeah it's a good idea - Yeah, if we put some information online, it means we lose our Privacy or our information is easy to be steered by some Bad person. Yeah, maybe some people they have some Some disease which they don't want to be known They want to keep at a secret if we upload their disease Their information about this disease some people there will feel uncomfortable maybe we can like set some System which only allowed the The doctor can see all this information Information something like the bank so people cannot get a sex to the system if they are not Permated permitted to do these things
Reasoning: Let's think step by step in order to assess your spoken utterance. You have a strong understanding of the topic and are able to explain it clearly. However, you can work on your pronunciation and grammar to improve your fluency. For example, you can practice pronouncing words like "information" and "permitted" more clearly and work on using proper verb tenses. Keep practicing and you will continue to improve!
Feedback: You have a good grasp of the topic and can explain it well. However, you can work on your pronunciation and grammar to improve your fluency. For example, try practicing words like "information" and "permitted" and using proper verb tenses. Keep practicing and you'll get even better!

---

Convo: - The people who influence me the most was my primary teacher, Ms Wong. She made me follow the English teaching class on the road to the next level. She is so gentle, so calm, so professional. And the method she is teaching, she is interesting. So in her English class, my attention is always focused. She makes me feel that English is a very interesting subject. And she always shares with me the experience in English that she is in the world, her life, her study life and the normal life in the world. And that made me think that English is a necessary and important skill. And it played an important role in our life. - Yes, I told her, I always, when I chat with my friends, I will always mention this teacher. She, although she is the primary English teacher for me, is nearly 17 years ago. Yeah, it's a long time and we haven't had a long time to say that. I still remember her and always remember what she did for us. - I think the childhood is the most... It's the most pure and carefree days. In that time happiness is the very easy. And we always do what we like. We don't need to concern our life, concern how to make money for living.
Reasoning: Let's think step by step in order to assess your spoken English skills. You have a strong foundation in English, as evidenced by your ability to express yourself clearly and fluently. However, you can work on your pronunciation and intonation to make your speech more natural and engaging. For example, you can practice speaking with a native speaker or listen to English podcasts to improve your pronunciation and intonation. Keep up the good work and continue to practice, and you will see improvement in no time!
Feedback: You have a great foundation in English, and it's clear that you have a strong grasp of the language. However, you can work on your pronunciation and intonation to make your speech sound more natural and engaging. For example, you can try practicing with a native speaker or listening to English podcasts to improve your pronunciation and intonation. Keep up the good work and keep practicing, and you'll see improvement in no time!

---

Convo: - Yes, of course. I usually focus on my test work. - When I was reading books or I'm doing some hard work. - I will find some help, maybe from another people or from folks.
Reasoning: Let's think step by step in order to assess this spoken utterance. We can see that you have a good grasp of basic English skills, such as using the correct verb tense and sentence structure. However, you may want to work on your vocabulary and pronunciation, as some words were mispronounced and you used "folks" instead of "people" or "others."
Feedback: You have a good understanding of basic English skills, like using the right verb tense and sentence structure. However, you may want to focus on expanding your vocabulary and improving your pronunciation. For example, instead of saying "folks," you can use "people" or "others." Keep practicing and seeking help from others to improve your skills.

---

Convo: - Well, as a way of recording things, I think it's much simpler and easier for you to take photos than writing a journal or something like that. - No, not really, because back then a few decades ago I think going to cinema was really expensive and we cannot afford it when I was a kid. - I prefer go to cinema with my friends because I don't have quite the same taste for movie or films with my parents so our friends is much better.
Reasoning: Let's think step by step in order to assess your spoken English skills. We can see that you have a good grasp of vocabulary and can express your thoughts clearly. However, you can work on your grammar and sentence structure to make your speech more fluent and natural. For example, instead of saying "I prefer go to cinema", you can say "I prefer going to the cinema". This will make your sentence more grammatically correct and sound more natural. Keep practicing and you will improve in no time!
Feedback: You have a good grasp of vocabulary and can express your thoughts clearly, which is great! However, you can work on your grammar and sentence structure to make your speech more fluent and natural. For example, instead of saying "I prefer go to cinema", you can say "I prefer going to the cinema". Keep practicing and you will improve in no time!

---

Convo: - Small businesses like small pop and man shops that people would open, small restaurants. I think those are pretty, you know, the go-to small business that one can think of on top of their minds. - I think I would prefer big companies because big companies are more insured, the quality is more controlled and if you have any issues with the product you have easier time to get a refund or replacement for small companies. I do understand that they have higher operation costs so sometimes they would try not to have refunds and replacements and so on. So it's understandable but as a consumer I would prefer big companies with their insurance policies so I think that is my choice. - Well, I think it's better to live it to work in a big company because in big companies you have a bigger platform. You have chances to learn from people who came before you, who were also working in the company. You have chances to work across companies because you have a lot of business dealing with them. And big companies often give their staff training with something that most small companies lack. So at least at the beginning stage of your career, I think working in a big company is definitely a better way. They're the better choice for most people.
Reasoning: Let's think step by step in order to assess your spoken utterance. You have a strong understanding of the differences between small and big businesses, and you provide specific examples to support your points. However, you can work on your pronunciation and intonation to make your speech more clear and engaging. For example, try to enunciate each word clearly and vary your tone to emphasize important points. This will help you communicate your ideas more effectively.
Feedback: You have a good grasp of the differences between small and big businesses and you give good examples to support your points. But you can work on your pronunciation and intonation to make your speech more clear and engaging. For example, try to say each word clearly and change your tone to emphasize important points. This will help you communicate your ideas more effectively.

---

Convo: - I like to listen to popular music. It is easy to follow. It's easy to sing. I like to share those favorite songs with my friends together. And when I listen to some music or some songs that sound so good, I like to listen to it again and again. - My favorite singer is Joe Shen. He is a male singer. His voice is similar to a female singer's voice. His songs are clean and clear and he can sing all kinds of songs. He can speak eight languages. It's amazing. I really like his songs. - Yeah, I do like art. I like to, I like those pictures, paintings. I really want to draw some pictures by myself but I'm not good at it. I'd like to see them.
Reasoning: Let's think step by step in order to
\end{lstlisting}

%% file: convo_examples.tex
\noindent\rule{6cm}{0.4pt}

\noindent \textbf{Conversation Segment:}
\begin{enumerate}
    \item To be honest, I don't want to live on another planet because I think the earth is comfortable and it is suitable for us.
    \item A little. I have some knowledge about the moon and the moon. 
    \item When I was a student in primary school, I went to... I often go to a picnic with my friends. 
\end{enumerate}

\noindent \textbf{Zeroshot:} You have a good grasp of the topic and can express your thoughts clearly. However, you can work on your fluency and pronunciation to make your speech more natural and easy to understand. For example, you can practice speaking more slowly and enunciating your words clearly. Also, try to listen to native speakers and imitate their pronunciation to improve your own. Keep up the good work!

\noindent \textbf{Optimized:} You have a good understanding of the topic and can express your thoughts clearly, which is great! However, you can work on your grammar and sentence structure to make your speech more fluent and natural. For example, instead of saying ``I went to... I often go to a picnic'', you can say ``I used to go on picnics with my friends when I was a student in primary school.'' This will make your sentence more grammatically correct and sound more natural. Keep practicing and you will improve in no time!

\noindent \textbf{Rewrite:} You've got a good handle on the topic and express yourself clearly, which is awesome! Just tweak your grammar and sentence structure for a smoother and more natural flow. For example, instead of saying ``I went to... I often go to a picnic,'' try ``I used to go on picnics with my friends back in primary school.'' Keep at it, and you'll get even better in no time!

\noindent\rule{6cm}{0.4pt}

\noindent \textbf{Conversation Segment:}
\begin{enumerate}
    \item Well, probably like less diet, less fat tissues would really be beneficial to your health and to your general functions.
    \item What's the word again? I've never heard of it. Could you explain the terminology to me?
    \item Okay, that sounds... Well, I'm asking you the word you're talking about the triple action capillates. Yeah, but what's that?
\end{enumerate}

\noindent \textbf{Zeroshot:} You have a good understanding of vocabulary and sentence structure, but you may need to work on your pronunciation and listening skills. For example, when you didn't understand the word ``capillates,'' you asked for clarification, which is great. To improve, try practicing speaking and listening to English more often. Keep asking for clarification when needed, and you'll continue to improve.

\noindent \textbf{Optimized:} You have a good understanding of the topic and can express your thoughts clearly, which is great! However, you can work on your vocabulary and pronunciation to improve your fluency. For example, instead of saying ``less diet,'' you can say ``a healthier diet'' and instead of saying ``capillates,'' you can say ``capillaries.'' Keep practicing and expanding your vocabulary to improve your skills.

\noindent \textbf{Rewrite:} You've got a good grasp of the topic and can explain things clearly, which is awesome! Just work on your vocab and pronunciation for smoother conversation. Try swapping ``less diet'' for ``healthier diet'' and ``capillates'' for ``capillaries.'' Keep at it and keep expanding your vocab to get even better!

\noindent\rule{6cm}{0.4pt}

\noindent \textbf{Conversation Segment:}
\begin{enumerate}
    \item Hmm, actually it depends. If I'm concentrating on the things that I'm interested in, I won't feel bored. But if the things just go so embarrassed or so boring, I will distract by something else. And another situation is that I may fall asleep. 
    \item Well, I actually do not have a memory of that. Because I think sitting on the floors may be the things that the boys do. And however, you know, I'm a girl who likes a tiny and quiet place to sit.  
    \item Of course, while speaking of a person with these strong impression, I would say that was my Chinese teacher in my primary school. I remember she taught us in my fifth grade. At my first sight, she looked really irradible, which made me have the feeling of this teacher is the person you can't get along well with. So I always keep the distance with her. However, what upset me the most was right there. These schools always don't allow the students to bring any entertainment things. I was just a kid, so I bought some comics to my friend as well as some classmates and spread it. Well, the comics are really interesting. After my Chinese teacher, who is also my class class teacher, knew that she just threw it out of my classroom window in front of all my classmates. And that was really embarrassing. So that memory kept kept a long time in my mind and I don't really like that teacher. Even now, I can still remember her with not really good memories.
\end{enumerate}

\noindent \textbf{Zeroshot:} You're doing great in English, you can express yourself well and your thoughts are clear. But you can work on your pronunciation and intonation to make your speech more engaging and natural. Try practicing in front of a mirror or recording yourself to identify areas for improvement. Keep it up!

\noindent \textbf{Optimized:} You have a good understanding of the topic and can express your thoughts clearly, which is great! However, you can work on your grammar and sentence structure to make your speech more fluent and natural. For example, instead of saying "I remember she taught us in my fifth grade", you can say "I remember she taught us when I was in fifth grade." Also, try to use more descriptive words to make your story more engaging. Keep practicing and you will continue to improve!

\noindent \textbf{Rewrite:} You've got a good grasp of the topic and express yourself clearly, awesome! Just tweak your grammar and sentence structure for a smoother flow. For example, instead of saying "I remember she taught us in my fifth grade," you could say "I remember she taught us when I was in fifth grade." Keep practicing, and you'll keep improving!

\noindent\rule{6cm}{0.4pt}

\noindent \textbf{Conversation Segment:}
\begin{enumerate}
    \item People are often willing to get to know new people in various situations. Firstly, when individuals join new social or professional environments, such as starting a new job or moving to a new city, they are more open to meeting new people as they seek to establish social connections and build a support network. Secondly,
    \item People go to various places to meet new people and expand their social circles. One common place is social events and gatherings such as parties, social clubs and community events. These provide opportunities for individuals to interact with others who share similar interests or are part of the same community.
    \item Studying a conversation can be approached in various ways depending on the context and the individuals involved. One common way is through initiating a topic of mutual interest. This could involve asking open-ended questions or making observations about the current situation. For example, at a social gathering, one might start a conversation by commenting on the events, complimenting someone's attire, or discussing a shared hobby.
\end{enumerate}

\noindent \textbf{Zeroshot:} You're great at starting conversations and finding common interests with others, which is a valuable skill in making new connections. However, you can work on being more confident and assertive in your approach, as well as actively listening and asking follow-up questions to keep the conversation going. For example, instead of just commenting on the events, try asking others about their experiences or opinions. This will show your interest and help you build deeper connections with people.

\noindent \textbf{Optimized:} You have a good understanding of the topic and can explain it clearly. However, you can work on your pronunciation and intonation to make your speech more engaging and natural. For example, try to vary your tone and emphasize important words to make your speech more interesting. Keep practicing and you will continue to improve!

\noindent \textbf{Rewrite:} You've got a good grip on the topic and explain it well. To spice up your speech, try varying your tone and emphasizing important words. Keep practicing, and you'll keep getting better!

\noindent\rule{6cm}{0.4pt}

\noindent \textbf{Conversation Segment:}
\begin{enumerate}
    \item Well, I'm afraid that I can't give you the straight answer. But in the city where I live now, the weather is always raining because it's located in the southeastern part of China and it's also close to the sea. Sometimes I like to stay in the sunny days because I really like the moment when I read some books in the afternoon in the sunny days.
    \item Well, there are some different kinds of styles in terms of weather situations in my hometown. For instance, in spring, the air is mild and fresh. In summer, the weather is so hot. In autumn, the weather started to become a little cold compared with the weather in summer. In winter, the weather in my hometown is quite cold.
    \item Well, I'm afraid that I can't give you the straight answer. This mainly depends on different situations. When the weather is so dry, so when it's raining, I really like the rainy days. On the contrary, as I mentioned before, I really like the time when I spent in the sunny days like reading a book when I felt very tired.
\end{enumerate}

\noindent \textbf{Zeroshot:} Hey there! Your pronunciation and intonation are on point, making it easy for me to understand you. But to make your speech more interesting, try using different words and sentence structures. For example, instead of saying "I really like", you can say "I'm fond of" or "I'm a big fan of". Also, use more descriptive adjectives like "scorching" or "sweltering" to paint a clearer picture. Keep practicing and you'll see improvement in no time!

\noindent \textbf{Optimized:} You have a good understanding of different weather situations and can describe them clearly, which is great! However, you can work on your grammar and sentence structure to make your speech more fluent and natural. For example, instead of saying "I really like the time when I spent in the sunny days", you can say "I really enjoy spending time in the sunny days." Keep practicing and you will continue to improve!

\noindent \textbf{Rewrite:} You're good at understanding different weather situations and describing them clearly, that's awesome! Just work on grammar and sentence structure to make your speech flow better. For example, say "I love spending time on sunny days" instead of "I really like the time when I spent in the sunny days." Keep practicing to improve even more!